\documentclass[]{article}
\usepackage{proceed2e}

\pdfoutput=1

\usepackage{graphicx} 
\usepackage{subfigure} 
\usepackage{amsmath}
\usepackage{amssymb}
\usepackage{natbib}

\usepackage{algorithm}
\usepackage{algorithmic}
\usepackage{comment}
\usepackage{bm}
\usepackage{hyperref}

\def\Dir{\mbox{Dirichlet}}
\def\GEM{\mbox{GEM}}
\newcommand{\hidetext}[1]{ }

\def\rest{\bm{rest}}

\newcommand\wrdtbl[1]{{\tau^{word}_{#1}}}
\newcommand\tmtbl[1]{{\tau^{time}_{#1}}}
\newcommand\wrddsh[1]{{d^{word}_{#1}}}
\newcommand\tmdsh[1]{{d^{time}_{#1}}}

\newcommand\allWD[1]{\bm{\wrddsh{#1}}}
\newcommand\allTD[1]{\bm{\tmdsh{#1}}}
\newcommand\allWT[1]{\bm{\wrdtbl{#1}}}
\newcommand\allTT[1]{\bm{\tmtbl{#1}}}

\title{A non-parametric mixture model for topic modeling over time} 

\author{ 
{\bf Avinava Dubey \quad Ahmed Hefny \quad Sinead Williamson \quad Eric P. Xing} \\
Machine Learning Dept. \\
Carnegie Mellon University
} 

\begin{document} 

\maketitle

\begin{abstract}
A single, stationary topic model such as latent Dirichlet allocation is inappropriate for modeling corpora that span long time periods, as the popularity of topics is likely to change over time. A number of models that incorporate time have been proposed, but in general they either exhibit limited forms of temporal variation, or require computationally expensive inference methods. In this paper we propose non-parametric Topics over Time (npTOT), a model for time-varying topics that allows an unbounded number of topics and flexible distribution over the temporal variations in those topics' popularity. We develop a collapsed Gibbs sampler for the proposed model and compare against existing models on synthetic and real document sets.
\end{abstract}

\section{Introduction}

Latent variable models, such as latent Dirichlet allocation \citep[LDA,][]{blei:03} and hierarchical Dirichlet processes \citep[HDP,][]{teh06}, are popular choices for modeling text corpora. Documents are modeled as a distribution over a shared set of topics, which are themselves distributions over words. Each word in a document is assumed to be generated by one of these topics.

Most topic models assume that the  documents are \emph{exchangeable}, or in other words, that the order in which they appear is irrelevant. This is often not a reasonable assumption -- the distribution over topics in today's newspaper is likely to be more similar to the distribution over topics in yesterday's newspaper than to the distribution over topics in a paper from a year ago. Similarly, popular topics on Twitter are likely to vary with both time and geographic location.

A number of models have been proposed to address this. Dependent
Dirichlet processes \citep{MacEachern:1999} are distributions over
collections of distributions, each indexed by a location in some
covariate space (e.g. time), such that distributions that are close
together in that space tend to be similar. Various forms of dependent
Dirichlet process have been used to construct time-dependent topic
models. Many of these models are limited in the form of variation
obtained -- for example the in the models of \citet{Lin_constructionof} and \citet{Rao:Teh:2009} the probability of seeing a topic as a function of time is restricted to be unimodal. Moreover, these models are difficult to apply to higher dimensional spaces, and often rely on the discretization of time. More flexible models, such as \citet{Srebro05time-varyingtopic} and \citet{MacEachern:2000}, tend to lose the desireable conjugace properties of the corresponding stationary model, making inference challenging.

An alternative approach is seen in a model known as Topics over Time \citep[TOT,][]{wang:06}. Unlike the previously discussed models, which define a distribution over topics conditional on a time, TOT models the text and the timestamp of a document jointly. This allows us to consider the timestamp as a random variable, rather than a fixed parameter. Such a framework allows us to incorporate non-Markovian dynamics while maintaining reasonable inference requirements. It also means that we can incorporate data without covariate information (for example, documents with no timestamp), something that is not easily achieved in conditional models such as dependent Dirichlet processes.

\begin{figure}
\label{fig:multimodal}
\includegraphics[width=0.45\textwidth,height=0.25\textheight]{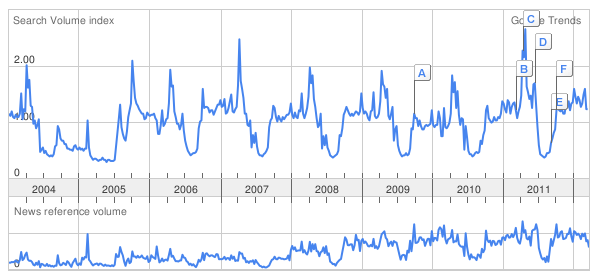}
\caption{Search and news trends for "NHL" obtained from Google Trends. This shows that interest in this topic rises and falls multiple times.}
\end{figure}

Like the conditional models, Topics over Time suffers from a number of shortcomings. The distribution over times for each topic is assumed to be unimodal, while in real life we often see topics vary in popularity in a more flexible manner. For example, Figure~\ref{fig:multimodal} shows the popularity of the search term ``NHL'' as a google query. The popularity waxes and wanes with the hockey season, and occasionally peaks due to a major news event. In addition, the number of topics must be fixed a priori, which can involve expensive model comparison.

In this paper, we propose a nonparametric extension to the Topics over Time model (npTOT). This model extends TOT to allow an unbounded number of topics, each of which can peak in popularity an unbounded number of times. In addition, npTOT induces correlations between the temporal variations in topic popularity, so that related topics trend in similar manners. Because, like TOT, npTOT is a joint model of both text and time, document/timestamp pairs can be considered exchangeable and we can make use of tractable exchangeable distributions to develop a Gibbs sampling scheme. We compare npTOT with its parametric counterpart, plus several baselines, and show that the added flexibility translates into qualitatively and quantitatively better performance.

\section{Related Work}
\label{sec:related}
Traditional topic models such as LDA have two main shortcomings. Firstly, they are parametric models that assume a fixed prespecified number of topics regardless of the data. Secondly, they assume that the probability of seeing a topic is independent of the time at which a document is written. In this section, we consider existing models that address one or both of these limitations.

\subsection{Nonparametric topic models}
To relax the assumption of a fixed number of topics, nonparametric topic models have been proposed. Rather than the fixed, finite number of topics specified by LDA, such models allow a countably infinite number of topics a priori, meaning that a random number of topics will be used to represent a given dataset. The most widely used nonparametric topic model replaces the collection of Dirichlet distributions used to model the per-document distributions over topics in LDA with a hierarchical Dirichlet process \citep[HDP,][]{teh06}. Here, the distribution over topics in a given document is given by a Dirichlet process. The document-specific Dirichlet processes are coupled using a shared base measure, which is itself a Dirichlet process.

\subsection{Dependent Dirichlet processes}

The HDP assumes that the documents in our corpus are exchangeable. A class of models referred to as dependent Dirichlet processes \citep[DDPs,][]{MacEachern:1999} relaxes this assumption. In topic models based on DDPs, each document is associated with a value in some covariate space, for example time. As in the HDP, the topic distribution of each document is marginally distributed according to a Dirichlet process. Unlike the HDP, documents that are close together in covariate space tend to have similar distributions. 

A number of DDPs have been used in topic modeling. The recurrant Chinese restaurant process \citep{ahmed:10} creates a Markov chain of distributions; however the model is non-exchangeable so we cannot make use of conjugacy in inferring the topic proportions. In addition, the model is only applicable to covariate spaces of a single dimension. A number of related models \citep{Caron_generalizedpolya, Lin_constructionof, Rao:Teh:2009} maintain some of the conjugacy of the original model, but do not allow as flexible variation in topic probability. A number of DDPs can exhibit more flexible, non-Markovian variation in topic probabilities \citep{Srebro05time-varyingtopic, MacEachern:2000}, but inference in such models scales very poorly.

\subsection{Topics over Time}

The DDP models mentioned in the previous section are examples of conditional models -- the covariate is assumed fixed, and the model defines a distribution over topics conditioned on this covariate value. The Topics over Time \citep[TOT,][]{wang:06} model takes a different tack, assuming that the covariate values are also random, and that the latent topics describe a distribution both over words and over times. This model is exchangeable if we consider a datapoint to consist of both a document's text and its timestamp, meaning we can make use of conjugacy. 

TOT is a form of supervised LDA \citep{Blei:McAuliffe:2007}, where the label is the timestamp of the document. TOT assumes the following generative process for a corpus of documents and their associated timestamps:

\begin{enumerate}
\item For each topic$k=1,\dots, K$
\begin{enumerate} 
  \item Sample a distribution over words $\phi_k \sim \Dir(H)$.
  \item Choose a set of parameters $\psi_k$ to parametrize a beta distribution.
\end{enumerate}
\item For each document $j=1,2,\dots,D$
  \begin{enumerate}
  \item Sample a distribution over topics, $\theta_j| \alpha \sim Dir(\alpha)$.
  \item For each word $i = 1,\dots, N_j$
    \begin{enumerate}
    \item Sample a topic indicator $z_{ji}|\theta_j\sim \theta_j$.
    \item Sample a word $w_{ji} | z_{ji} \sim \mbox{Mult}(\phi_{z_{ji}})$.
    \item Sample a timestamp $t_{ji} | z_{ji} \sim \mbox{Beta}(\psi_{z_{ji}})$.
    \end{enumerate}
  \end{enumerate}
\end{enumerate}

This model exhibits non-Markovian variations in topic probabilities, but has a number of drawbacks. The beta distribution used to model the time-varying probability is unimodal, and means that times must be bounded. This limits the form of temporal variation available, and precludes prediction outside of the bounded time-frame or extension to higher dimensionalities. Moreover, the lack of a prior on $\psi_k$ means it must be estimated using an approximate method.  In addition, the number of topics must be defined a priori.

\section{Nonparametric Topic Over Time (npTOT)}
\label{sec:nptot}

In this section we address the two problems identified in ToT: Inflexible topic probability variation, and a fixed number of topics. The resulting model employs nonparametric distributions to generate both the distribution over topics, and the distribution over timestamps; therefore, we refer to this model as nonparametric Topics over Time (npTOT).

We follow TOT in assuming that each document (indexed by $j$) consists of a (unordered) set of tokens (indexed by $i$). Each token $x_{ji} := (w_{ji}, t_{ji})$ is defined to be an ordered pair of a word $w_{ji}$ and a timestamp $t_{ji}$ \footnote{In practice, a document has a single timestamp which we duplicate for each word during inference.}. We assume that each document is generated by a distribution over multiple topics. 

The restriction to a fixed number of topics can be avoided by replacing the Dirichlet distribution over topics with a hierarchical Dirichlet process. This allows an unbounded number of topics a posteriori, and ensures the topics are shared across documents.

The form of temporal variation can be modified by replacing the beta distribution in the ToT model with another choice of distribution. In order to model multimodal variation on an unbounded timeframe, while maintaining tractable inference, we choose to use a mixture of Gaussians. To allow a flexible distribution over time, we use a Dirichlet process as the mixing measure.

We note that there may be correlations between the trending patterns of topics, something that is not addressed in much of the dynamic topic modeling literature. For example, topics to do with a sports players and sports fans are likely to have similar temporal variation. We address this by allowing the components of our mixture of Gaussians to be shared between topics. This is achieved by sampling the mixture components from a hierarchical Dirichlet process. 

\begin{figure}[tb]
\centering
\includegraphics[width=.9\columnwidth]{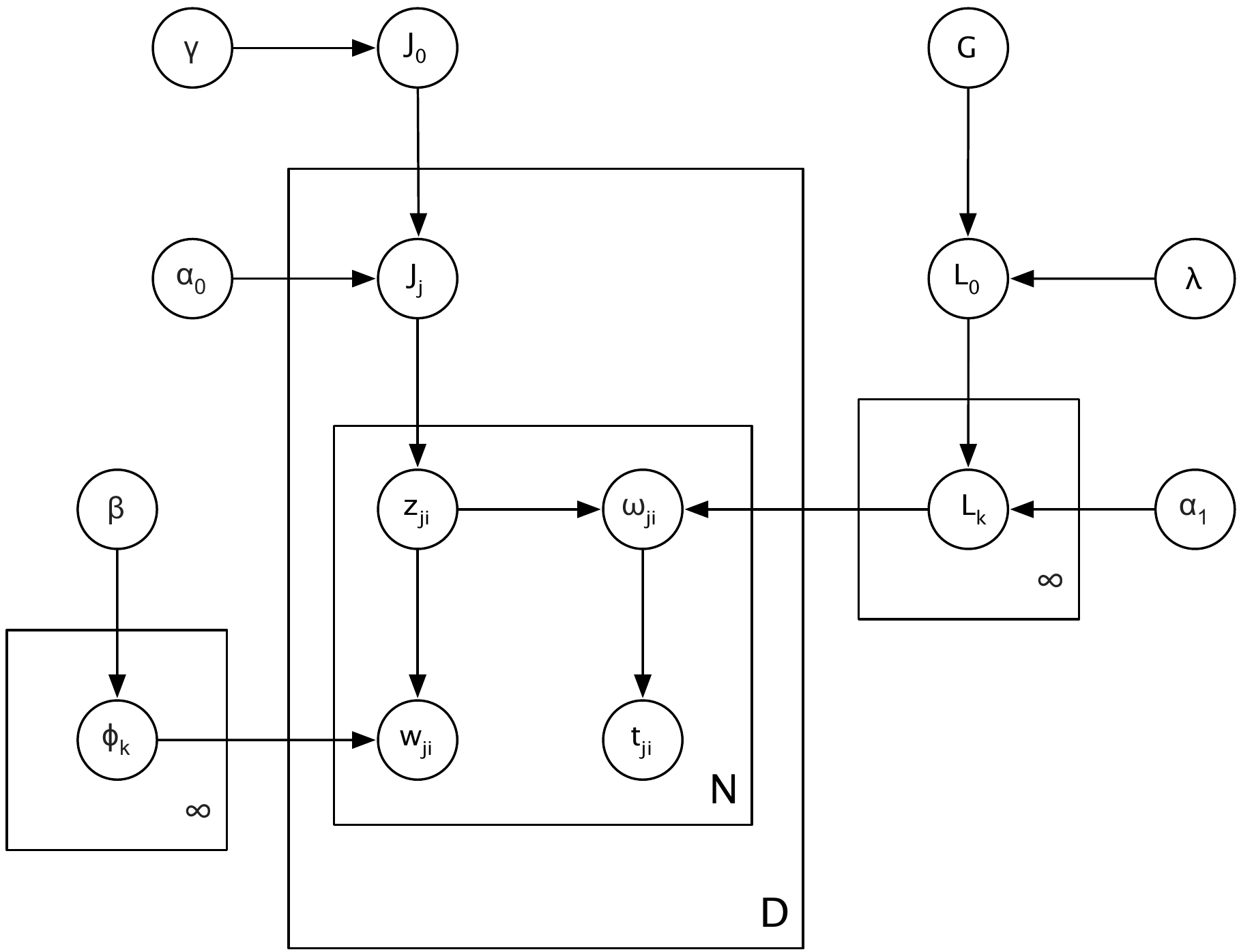}
\caption{Plate diagram for nonparametric Topics over Time}
\label{fig:model}
\end{figure}

Let $G$ be a normal-inverse Gamma distribution over the mean and variance of our time components, and assume $\gamma$, $\alpha_0$, $\alpha_1$ and $\lambda$ to be fixed hyperparameters. Let GEM indicate the distribution over probability measures associated with the Dirichlet process. The generative process, represented by plate diagram \ref{fig:model}, is defined as follows:
\begin{enumerate}
\item Sample a global base distribution over topic proportions, $J_0 | \gamma \sim \GEM(\gamma)$.
\item Sample a global base measure over the means and variances of the time components, $L_0 |\lambda, G \sim DP(\lambda, G)$, such that each component parametrizes a univariate Gaussian.
\item For each topic $k = 1,2,\dots$, 
  \begin{enumerate}
  \item Sample a distribution over words, $\phi_k \sim \Dir(\beta)$.
  \item Sample a topic-specific distribution over time components, $L_k |\alpha_1, L_0 \sim DP(\alpha_1, L_0) $
  \end{enumerate}
\item For each document $j=1,2,\dots,D$
  \begin{enumerate}
  \item Sample a distribution over topics, $J_j| \alpha_0, J_0 \sim DP(\alpha_0,J_0)$
  \item For each word $i = 1,\dots, N_j$
    \begin{enumerate}
    \item Sample a topic indicator $z_{ji}|J_j\sim J_j$.
    \item Sample a word $w_{ji} | \phi_{z_{ji}} \sim \mbox{Mult}(\phi_{z_{ji}})$
    \item Sample a time component $\omega_{ji}:=(\mu_{ji},\sigma_{ji}) | L_{z_{ji}} \sim L_{z_{ji}}$
    \item Sample a timestamp $t_{ji} | \mu_{ji},\sigma_{ji} \sim \mathcal{N}(\mu_{ji},\sigma_{ji})$
\end{enumerate}
\end{enumerate}
\end{enumerate}


\section{Inference}
\label{sec:inference}

We propose a Gibbs sampler based on the Chinese restaurant franchise \citep[CRF,][]{teh06}. Our model requires {\em two} restaurant franchises, one for the word HDP and the other for the time HDP.

The CRF associated with the word HDP mimics that described by \citet{teh06}: each restaurant corresponds to a document and each dish corresponds to a topic. The CRF associated with the time HDP has a different interpretation: the restaurants correspond to the topics, and the dishes correspond to ``time components'', which are associated with a Gaussian distribution over time. We terms such as ``time table'', ``time dish'', ``word table'', and ``word dish'' to distinguish between the two franchises.

Each token $x_{ji} := (w_{ji}, t_{ji})$ is associated with a word table $\wrdtbl{ji}$ and a time table $\tmtbl{ji}$.  Each word table $a$ in document $j$ is associated with a word dish (topic) $\wrddsh{ja}$. Each time table $b$ in a topic $k$ is associated with a time dish (time component) $\tmdsh{kb}$. We define $n_{ja}$ as the number of tokens in document $j$ associated with word table $a$; $q_{kb}$ as the number of tokens associated with topic $k$ and time table $b$; and $f(v,k)$ as the number of times the word $v$ is associated with topic $k$.

At each iteration of our Gibbs sampler, we need to sample, for each token $i$ in document $j$, both the corresponding word table $\wrdtbl{ji}$ and time table $\tmtbl{ji}$. We also need to sample the topic $\wrddsh{ja}$ corresponding to each word table $a$ in document $j$ and the time component $\tmdsh{kb}$ corresponding to each time table $b$ in topic $k$. We describe these steps in detail in the remainder of this section.

\subsection{Sampling $\wrdtbl{ji}$}
Recall that each word table is associated not just with a distribution over words, but also with a distribution over time tables. If we were to sample the word table for a token conditioned on that token's time table, our sampler would mix very slowly. Instead, we marginalize over $\tmtbl{ji}$ in order to sample $\wrdtbl{ji}$, and then sample $\tmtbl{ji}$ condtioned on $\wrdtbl{ji}$ as described in Section~\ref{sec:timesampling}.

The resulting distribution over time tables is given by

\begin{equation}
\begin{split}
&p(\wrdtbl{ji} = a | \bm{x}, \bm{t}, \rest_{-ji}) \\
\propto& p(\wrdtbl{ji} = a |  \allWT{-ji})\\
&p(w_{ji}, t_{ji} | \wrdtbl{ji} = a, \bm{x}_{-ji}, \bm{t}_{-ji},\rest_{-ji})\label{eq:wtbl}
\end{split}
\end{equation}
where $\rest_{-ji} = (\allWD{}, \allTD{}, \allWT{-ji}, \allTT{-ji})$. The component terms of Equation~\ref{eq:wtbl} are given by
\begin{equation}
p(\wrdtbl{ji} = a | \allWT{-ji}) \propto 
\begin{cases}
n_{ja}^{-ji} &\mbox{if $a$ is an existing table}\\
\alpha_0 &\mbox{otherwise,}\\
\end{cases}
\end{equation}
and

\begin{equation*}
\begin{split}
&p(w_{ji}, t_{ji} | \wrdtbl{ji} = a, \bm{x}_{-ji}, \bm{t}_{-ji},\rest_{-ji})\\
=& \frac{(\beta + f(w_{ji}, k))}{V\beta + \sum_{v=1}^V f(v,k)}\\
&p(t_{ji} | \wrdtbl{ji} = a , \wrddsh{ja} = k , \bm{t}_{-ji} , \rest_{-ji})
\end{split}
\end{equation*}
if $a$ is an existing table serving dish $k$, or
\begin{equation*}
\begin{split}
&p(w_{ji}, t_{ji} | \wrdtbl{ji} = a, \bm{x}_{-ji}, \bm{t}_{-ji},\rest_{-ji})\\
=&\sum_{k=1}^{K} \frac{m_{.k}}{m_{..} + \gamma} \frac{\beta + f(w_{ji}, k)}{V\beta + \sum_{v=1}^V f(v,k)}  \\
&p(t_{ji} | \wrdtbl{ji} = a , \wrddsh{ja} = k, \bm{t}_{-ji} ,\rest_{-ji}) \\ 
&+ \frac{\gamma}{m_{..} + \gamma} \frac{1}{V}\\
&p(t_{ji} | \wrdtbl{ji} = a , \wrddsh{ja} = k_{new}, \bm{t}_{-ji} ,\rest_{-ji}) 
\end{split}
\end{equation*}
if $a$ is a new table. In both cases, we can write
\begin{equation*}
\begin{split}
&p(t_{ji} | \wrdtbl{ji} = a , \wrddsh{ja} = k , \bm{t}_{-ji} , \rest_{-ji}) \\
=&\bigg\{ \sum_{b} \frac{q_{kb}^{-ji}}{q_{k.}^{-ji} + \alpha_1} g_{\tmdsh{kb}}^{\bm{t}_{-ji}}(t_{ji}) \\
& + \frac{\alpha_1}{q_{k.}^{-ji} + \alpha_1} \bigg( \sum_{c=1}^{C} \frac{r_{.c}}{r.. + \lambda } g_{c}^{\bm{t}_{-ji}}(t_{ji})\\
& \phantom{\mbox{spaaaaace}}+ \frac{\lambda }{r.. + \lambda } g_{c_{new}}(t_{ji}) \bigg)\bigg\}\, ,
\end{split}
\end{equation*}
where  $g_c^{\bm{t}_{-ji}}$ denotes the posterior predictive time distribution for token $x_{ji}$ conditioned on a time component $c$ and other timestamps associated with that time component. For the Gaussian model described here, the posterior predictive distribution is a t-distribution.

If a new word table is created for token $x_{ji}$ then we sample its corresponding word dish (topic) from the global word DP. 

\subsection{Sampling $\wrddsh{ja}$}
In order to resample the topic assignment $\wrddsh{ja}$ for an entire word table, we need to marginalize over the time table assignments of \emph{all} the tokens (denoted $\bm{x}_{ja} = (\bm{w}_{ja}, \bm{t}_{ja})$) associated with that word table. Since the number of tokens at the word table might be large, summing over all possible assignments is infeasible, so we approximate $p(\bm{t}_{ja} | \wrddsh{ja} = k, \bm{t}_{-ja}, \rest_{-ja})$ by sampling sets of table topic assignments. We use the resulting estimate $\hat{p}(\bm{t}_{ja})$ to approximate the true Gibbs sampling probabilities:

\begin{equation*}
\begin{split}
&p(\wrddsh{ja} = k|\allWD{-ja}, \bm{w},\rest_{-ja})\\
\propto&
\begin{cases} m_{.k}^{-ja}p(\bm{w}_{ja} | \wrddsh{ja} = k, \bm{w}_{-ja}) \hat{p}(\bm{t}_{ja}) & \mbox{if existing topic}\\
\gamma p(\bm{w}_{ja} | \wrddsh{ja} = k, \bm{w}_{-ja}) \hat{p}(\bm{t}_{ja}) & \mbox{otherwise.}
\end{cases}
\end{split}
\end{equation*}
\subsection{Sampling $\tmtbl{ji}$ and $\tmdsh{kb}$}\label{sec:timesampling}
Given the topic assignments, the distribution over the timestamps is
independent of the rest of the model, and we can perform Gibbs
sampling as in \cite{teh06}.



%
\section{Evaluation}
The goal of this paper was to increase the flexibility of TOT, an existing joint model for documents and their timestamps, by allowing multimodal variation in topic popularity, and by learning the number of topics. In this section, we present experimental results that demonstrate that we can capture more flexible variation than TOT, and learn an appropriate number of topic components. Moreover, we show that this added flexibility translates into improved log likelihood on test datasets.
\subsection{Evaluation on Synthetic Data}
\begin{figure*}
\centering
\subfigure[]{\label{fig:Synthetic_a} \includegraphics[width=.9\columnwidth,height=0.08\textheight]{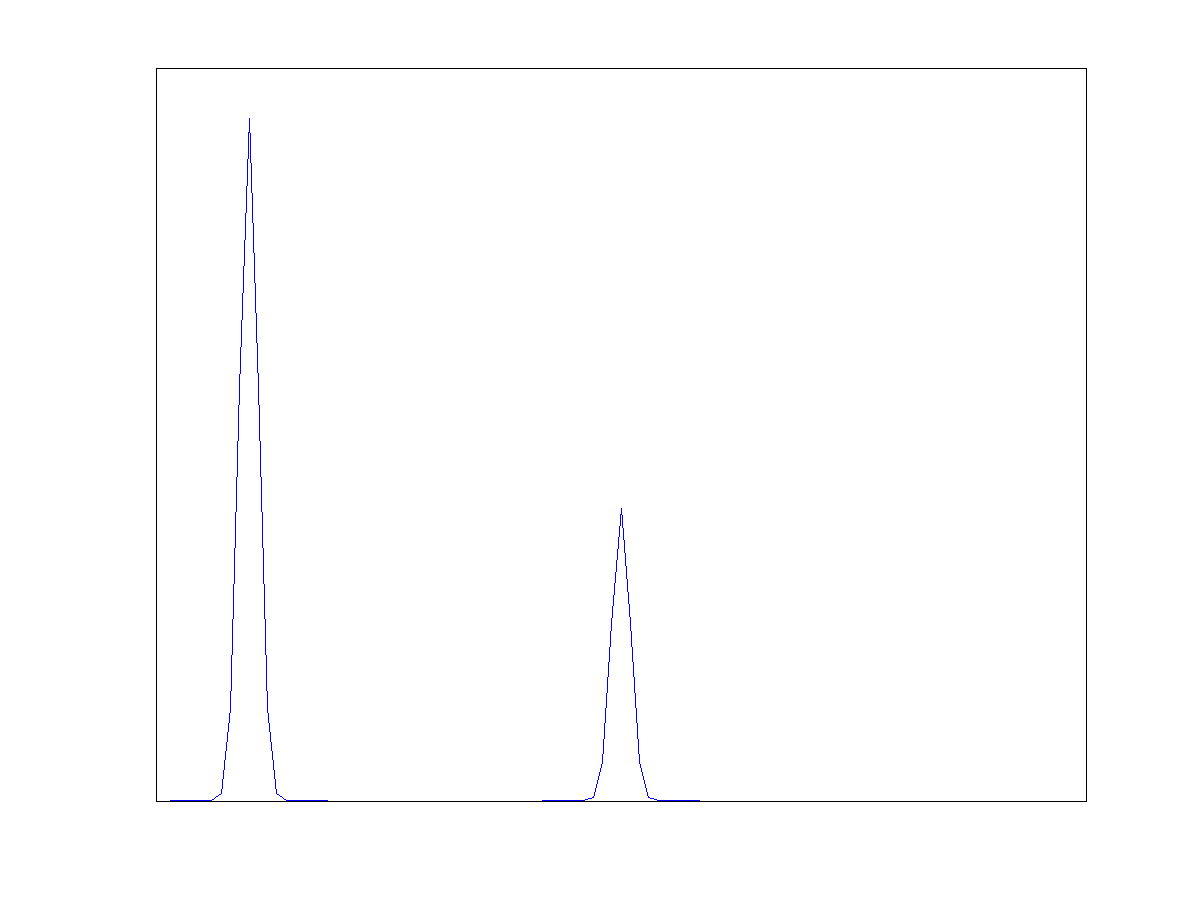}}
\subfigure[]{\label{fig:Synthetic_b} \includegraphics[width=.9\columnwidth,height=0.08\textheight]{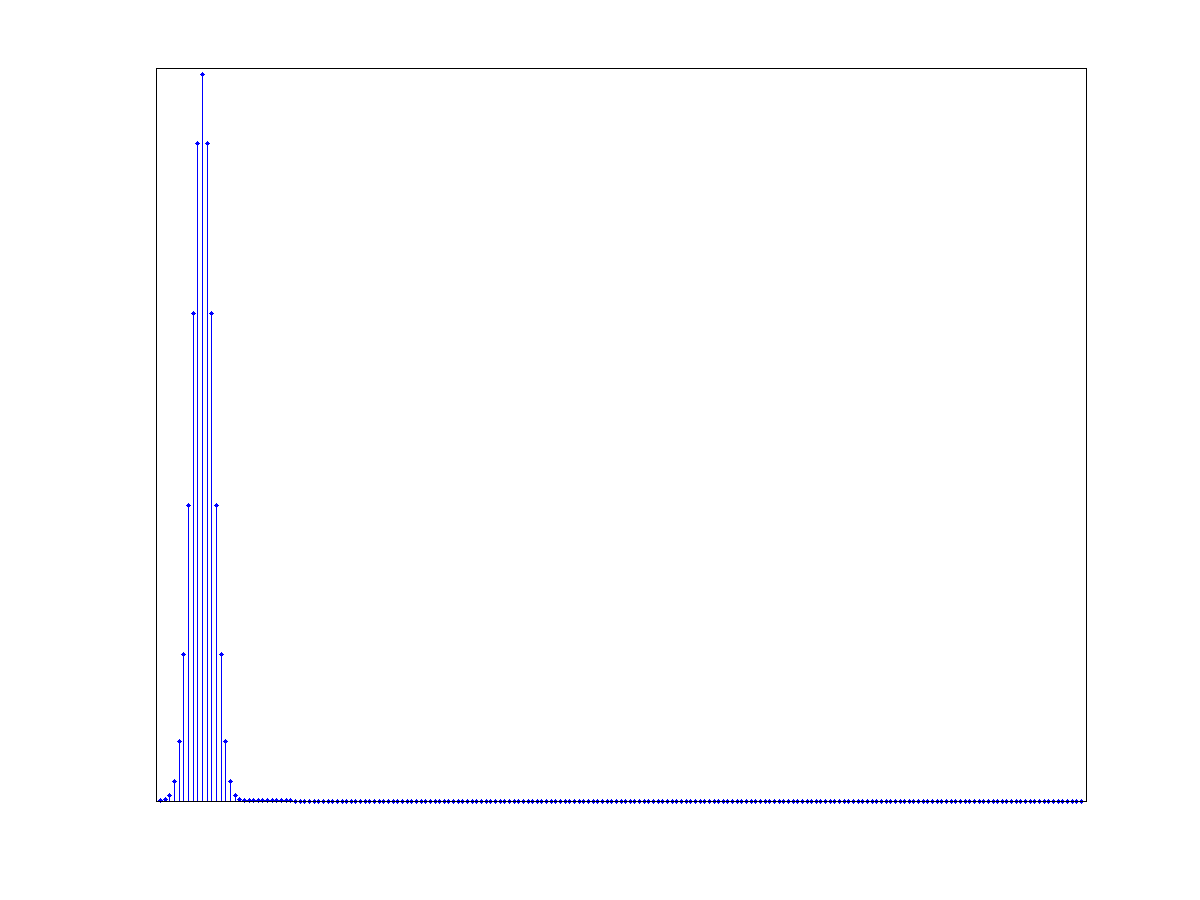}}
\subfigure[]{\label{fig:Synthetic_c} \includegraphics[width=.9\columnwidth,height=0.08\textheight]{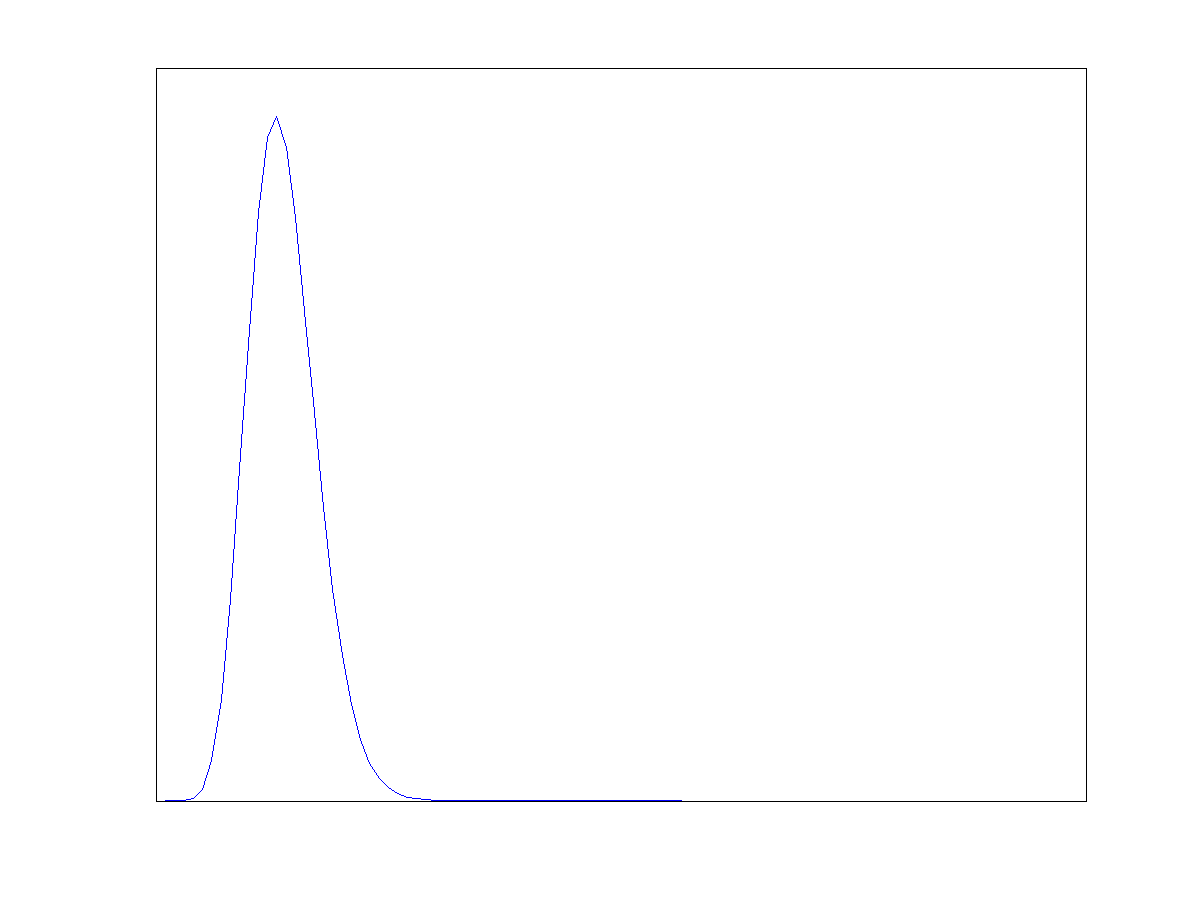}}
\subfigure[]{\label{fig:Synthetic_d} \includegraphics[width=.9\columnwidth,height=0.08\textheight]{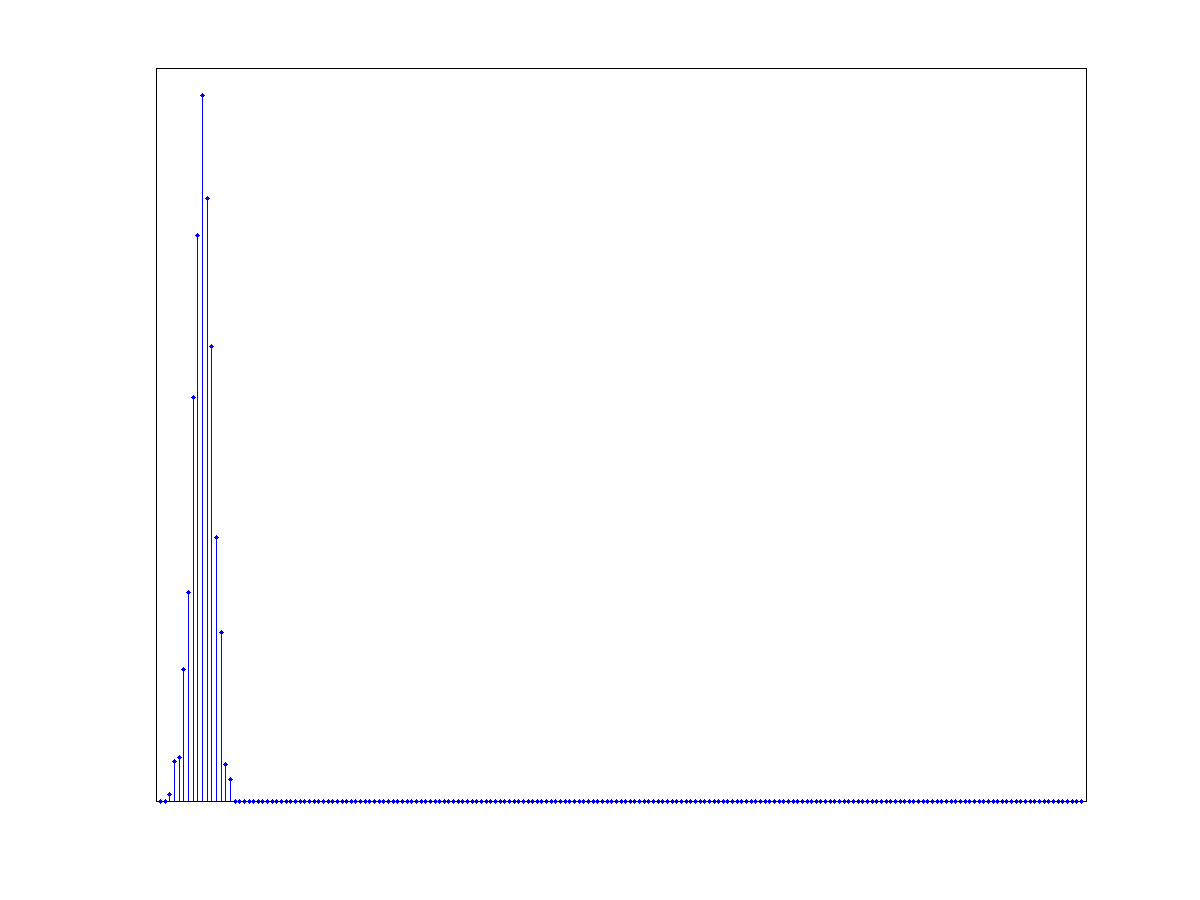}}
\subfigure[]{\label{fig:Synthetic_e} \includegraphics[width=.9\columnwidth,height=0.08\textheight]{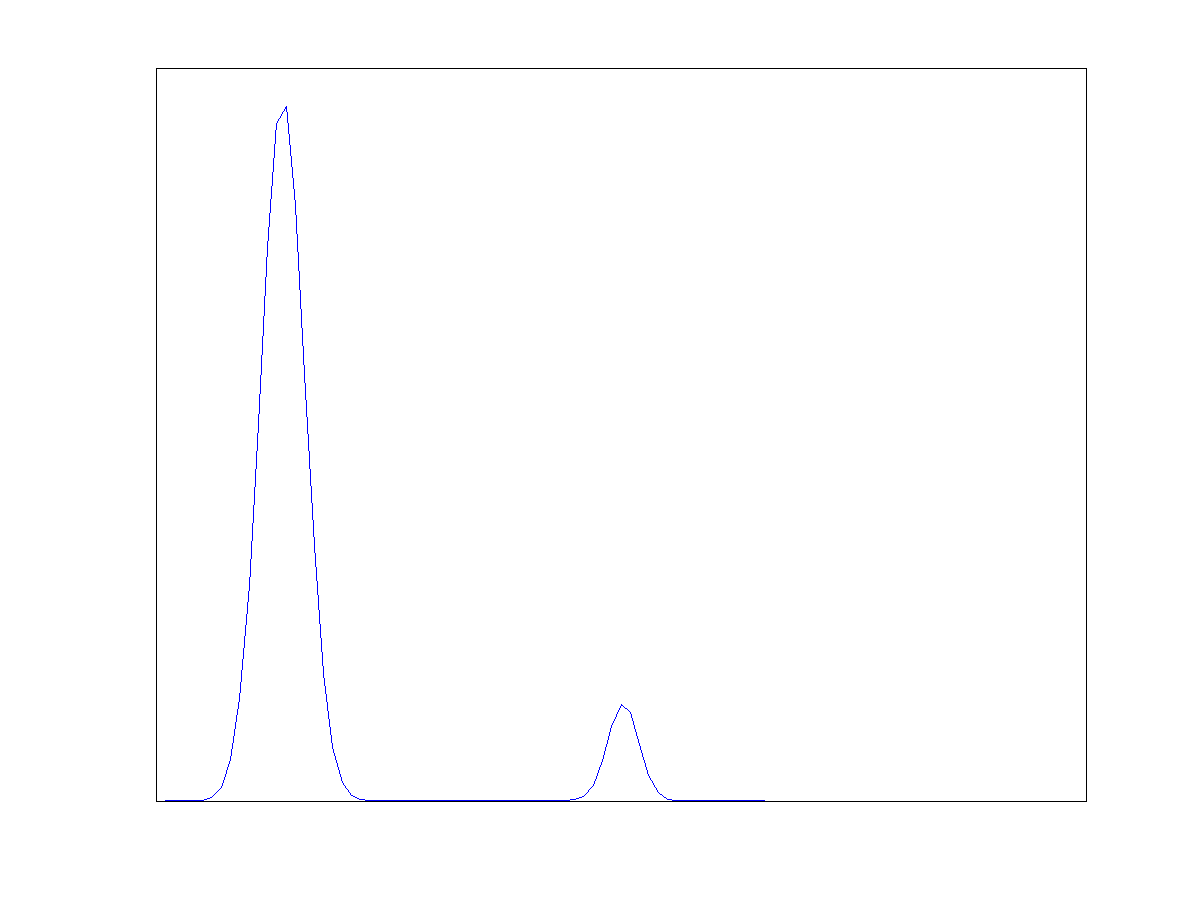}}
\subfigure[]{\label{fig:Synthetic_f} \includegraphics[width=.9\columnwidth,height=0.08\textheight]{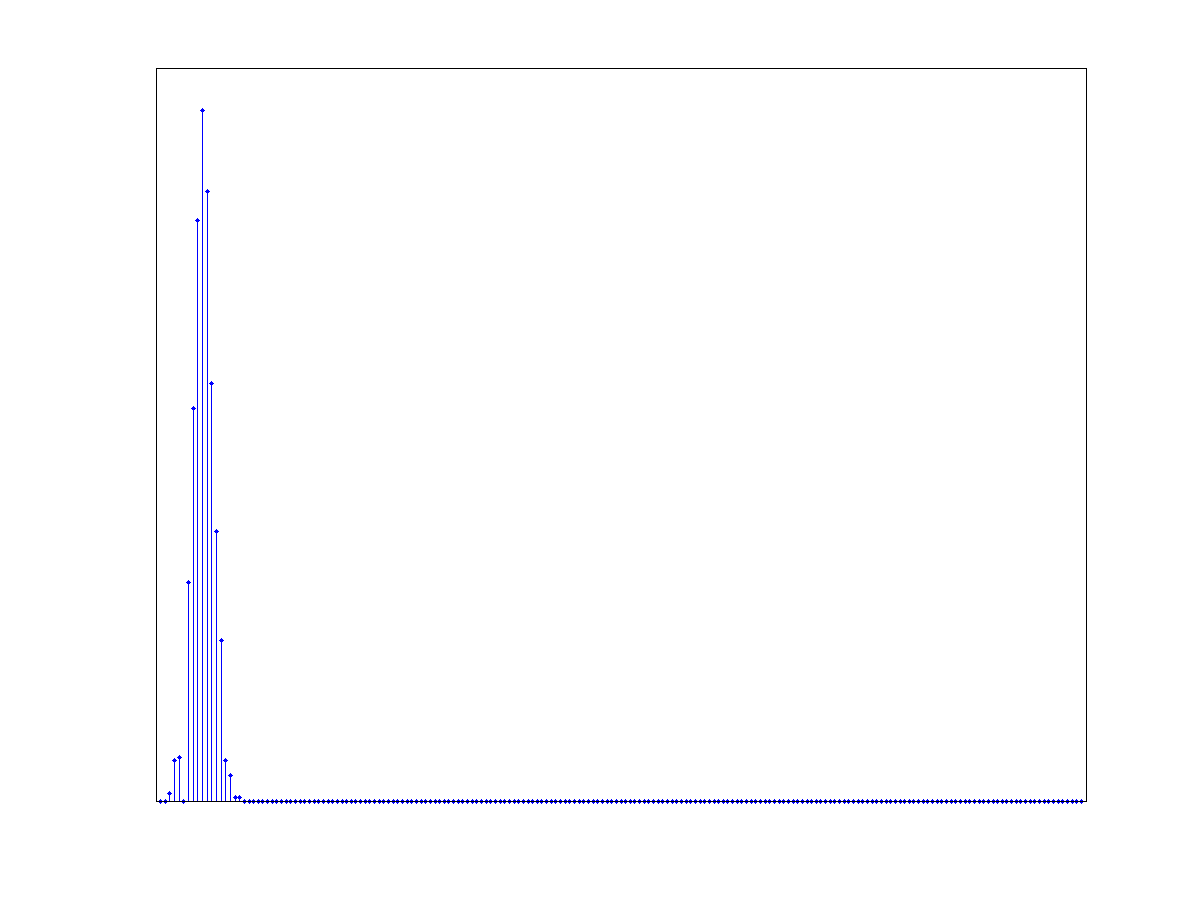}}
\caption{(a) shows the actual distribution over time for a particular topic on the synthetic data set, (b) shows the distribution over words for that particular topic, (c) shows the distribution over time of the TOT-detected topic closest to the original (d) shows the probability of the top-10 words for the TOT topics in (c), (e) shows the distribution over time for corresponding topic found by npTOT and (f) shows the word distribution for the npTOT topic in (e)}
\label{fig:Synthetic}
\end{figure*}

To demonstrate the ability of npTOT to recover the temporal variation of topics, we trained the model on a synthetic dataset, where ground truth is available. We generated a dataset of $D$ documents from $K$ topics, each associated with a multinomial distribution over $V$ words obtained by discretizing Gaussian distributions with means sampled uniformly on $[0,V]$. Each topic is also associated with a continuous distribution over time, distributed according to a mixture of $C$ Gaussians with means at $0.5 + k/K$, $k=1,\dots, K$. Each component has equal variance $\sigma$ such that $3*C*\sqrt{\sigma} = 1$. For each timestamp we generate one document. Each document is associated with a distribution over topics which is proportional to the probability of that document generating the timestamp. Topics and words were sampled according to the LDA generative procedure. We set $D$ to 100, $K$ to 30, $V$ to 100 and $C$ to 10. An example of a single topic and the corresponding distribution over times is shown in Figure \ref{fig:Synthetic}. 

We trained TOT and npTOT on the generated data. The number of topics
in TOT was set to the true number of topics, and as we see in Figure
\ref{fig:Synthetic}(b,d,e), the distributions over words obtained were
a good match for the generating data. The npTOT model found 27 topics,
very close to the true value. As we see in Figure~\ref{fig:Synthetic_c}, TOT was unable to capture the variation in topics. Conversely, npTOT was able to capture the multimodality of their distribution with respect to time (Figure \ref{fig:Synthetic_e}).
 
\subsection{Real-world Data Experiments}
To show that npTOT is able to capture the temporal variation in real documents, we performed experiments on three datasets:
\begin{itemize}
\item \textbf{Twitter Subset.} This dataset consists of tweets originating from Egypt in the time period from January through March 2011. We selected tweets given by active users where an active user is a user who has more than 200 tweets. As preprocessing, we removed words that are less than 3 characters long. We then removed the most frequent 40 words as well as words that occurred less than 10 times. Finally, we aggregate the tweets of each user in each day in a single document and remove documents that are less than 20 words long. The preprocessed dataset contains 6,072 documents, 9,080 unique words and 324,298 word tokens in total. Because these tweets originated from Egypt, they contain both Arabic and English words.
\item \textbf{State of the Union Address dataset.} The State of the Union dataset \footnote{http://www.gutenberg.org/dirs/text04/suall11.txt} contains the transcripts of 208 State of the Union addresses from 1790 to 2002. We followed \cite{wang:06} in processing the dataset. Namely, we divided each speech into three-paragraph documents, and removed stop words and numbers. This resulted in 5,897 documents, 22,620 unique words and 800,399 word tokens in total. 
\item \textbf{NIPS dataset.} The NIPS data set \footnote{http://cs.nyu.edu/~roweis/data.html} consists of the full text of the 12 years of proceedings from
1987 to 1999 Neural Information Processing Systems (NIPS) Conferences. The dataset is already preprocessed as described in \cite{chechik2007eec} and it consists of 1,740 research papers, 13,946 unique words and 2,301,375 word tokens in total.
\end{itemize}

In addition to TOT, which we will refer to as TOT-Unimodal in this section, we evaluated against three other baselines:

\begin{itemize}
\item \textbf{LDA-Unimodal} Here we ran LDA on the text of the documents, and then fit the temporal variation of each topic with a single Gaussian distribution.
\item \textbf{LDA-Multimodal} Here we ran LDA on the text of the documents, and then fit the temporal variation of each topic with a mixture of Gaussians.
\item \textbf{TOT-Multimodal.} Here, we restricted npTOT to have a fixed number of topics, in order to disambiguate the effect of an unbounded number of topics from the effect of using a more flexible distribution over time.
\end{itemize}

\begin{figure*}
\centering
\begin{tabular}{cc}
\includegraphics[width = .4\textwidth,height=0.15\textheight]{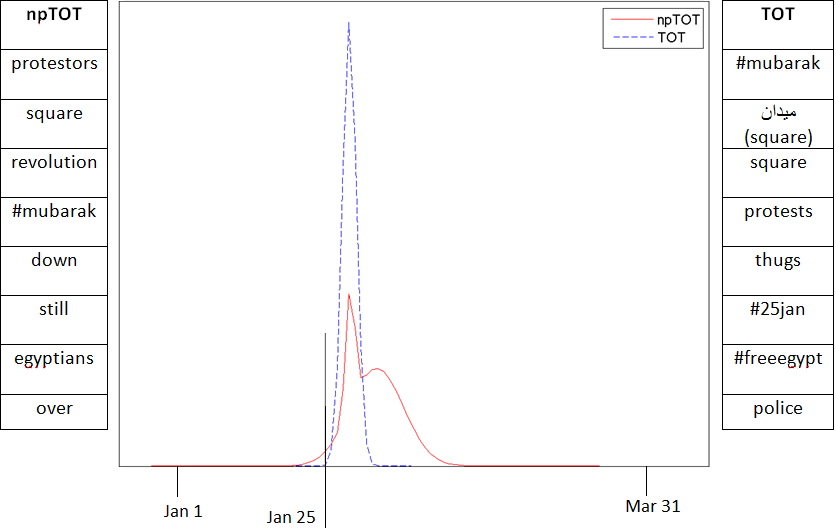} &
\includegraphics[width = .45\textwidth,height=0.15\textheight]{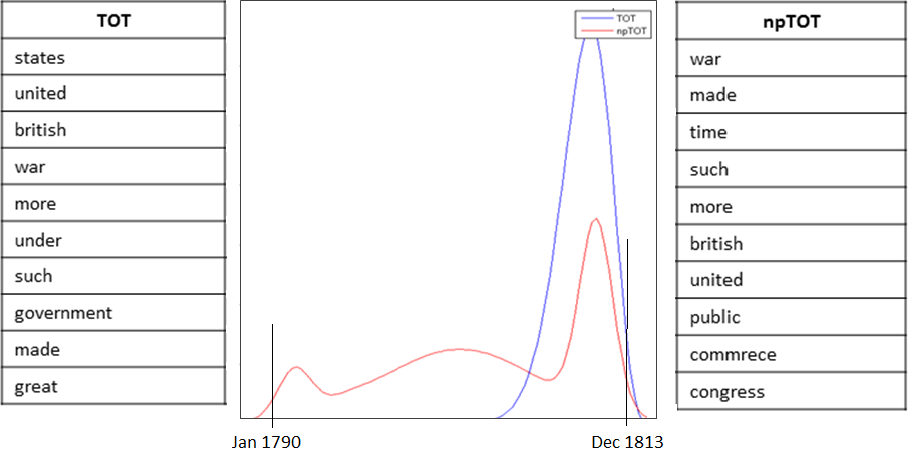}
\end{tabular}
\caption{\textbf{Left:} Top eight most probable words in topics from TOT and npTOT corresponding to the Egyptian revolution (started on Jan 25) in Twitter dataset.
\textbf{Right:} Top ten most probable words in topics from TOT and npTOT corresponding to conflicts involving the US and Britain in the State of the Union address dataset.}
\label{fig:twtTopic}
\end{figure*}

\subsection{Real-world Data Experiments}
To show that npTOT is able to capture the temporal variation in these datasets, we performed a qualitative analysis of the topics found, and a quantitative analysis of the predictive performance of the models.

\subsubsection{Qualitative analysis}
To see how npTOT can capture a wider variety of temporal variation than TOT, consider topics found using both models.
Figure \ref{fig:twtTopic} shows topics found in the Twitter and the State of the Union addresses. We hand-picked topics that addressed the same themes for the purpose of this comparison. On the Twitter dataset, we see a topic that arises with the outbreak of revolution in Egypt on January, 25, 2011.
Both models capture a sharp peak in this topic at that time, but the slow decay shown by the npTOT model is more realistic than the sharp decline in interest implied by the TOT model.  On a subset of the State of the Union dataset, we show a topic concerned with conflict involving the US and Britain. Both models show a sharp peak in this topic around the time of the War of 1812, but the nonparametric model is able to reuse this topic to describe tensions between the US and Britain leading to the declaration of the war, such as the Embargo Act of 1807. 

\begin{figure*}
\centering
\includegraphics[width = .9\textwidth,height=0.3\textheight]{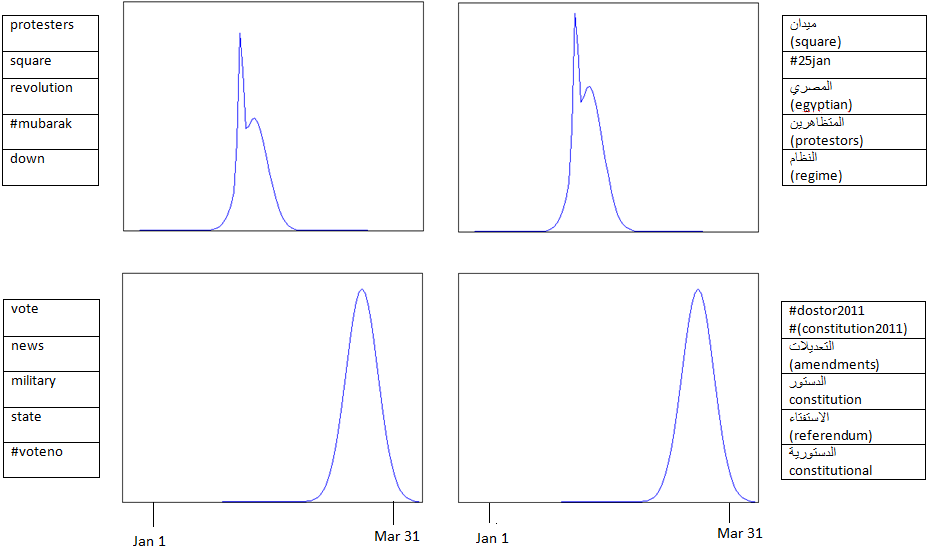}
\caption{Distributions over time for two discovered English topics (left) and their Arabic counterparts (right), showing that they share time components.The top topic is about the Egyptian revolution outbreak. The bottom topic is about a referendum on constitutional amendments. Each panel shows words selected from the top twelve most probable words in the corresponding topic. Words in parentheses are translated from Arabic.} 
\label{fig:time_cor}
\end{figure*}

Figure \ref{fig:time_cor} shows how related topics can share time components to give similar temporal variation. Since Twitter data is bilingual, we expect pairs of topics that address similar issues but in different languages. The figure shows two such pairs, demonstrating that they share the same time components.


\subsubsection{Quantitative analysis}
We evaluated the performance of npTOT and its competitors using two
methods: Joint likelihood of a document and its timestamp, and
perplexity of the second half of a document, conditioned on its
timestamp and the first half of the text. The joint likelihood gives a
general measure of how well the various methods are able to model the
corpora. The perplexity task demonstrates how well we are able
to make use of temporal information to predict the content of a
document. 

In each case, we randomly split each data set  into training and test
sets using a 70:30 split, and learned all four models on the training
set. The LDA models were run for 1000 iterations, and npTOT and TOT
were run until the percentage of changed tokens was below 5\%. The
joint log likelihood was obtained using the harmonic mean method, as
described in \cite{wallach:09}, by sampling topic assignments
$\bm{z}^{(s)}_d \sim \bm{z}_d|\bm{\Phi}, \bm{w}_d, \bm{t}_d$, (where
$\bm{\Phi}$ denotes the estimated model parameters) and taking the
harmonic mean of the conditional likelihoods $P(\bm{w}_d, t_d | \bm{z}^{(s)}_d, \bm{\Phi})$
over 200 samples.

We evaluated perplexity using the estimated $\theta$ method described in 
\cite{wallach:09}: for each test document $d$ we sample topic assignments $\bm{z}_d \sim \bm{z}_d|\bm{\Phi}, \bm{w}^{(1)}_d, \bm{t}_d$, where $\bm{\Phi}$ denotes the estimated model parameters, $t_d$ and $\bm{w}^{(1)}_d$ denote time stamp of document $d$ and the words in the first half of the document respectively.
 For each sample of $\bm{z}^{(s)}_d$ we estimate $\hat{\theta}^{(s)}_{dk} = P(k|\bm{z}^{(s)}_d, \alpha)$,
which we use to estimate the likelihood of the second half of the
document $P(\bm{w}^{(2)}_d|\bm{w}^{(1)}_d, \bm{\Phi},
\hat{\bm{\theta}}^{(s)}) = \prod_{i=1}^{N_d} \sum_k P(w_{di}|z_i = k,
\bm{\Phi}) \hat{\theta}^{(s)}_{dk}$. Taking the product over all
document and then averaging over samples of $\bm{z}^{(s)}$ gives an
estimate of the document completion likelihood. The perplexity score we report is evaluated as $\exp(-\mbox{completion log likelihood/N})$, where $N$ is the total number of words in the test set. 

\begin{figure}[]
\centering
\includegraphics[scale = .6]{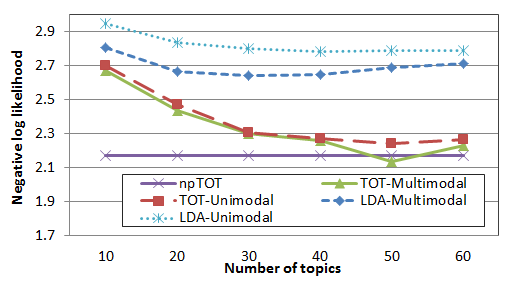}
\caption{Average per-token negative log likelihood on test set for Twitter dataset}
\label{fig:TLk}
\end{figure}
\begin{figure}[]
\centering
\includegraphics[scale = .6]{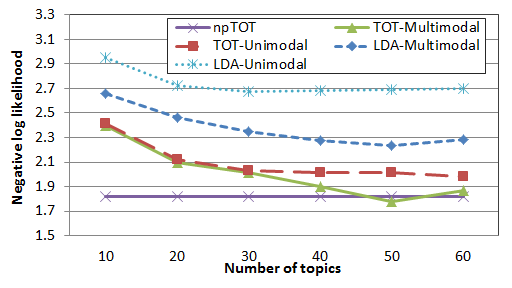}
\caption{Average per-token negative log likelihood on test set for State-of-the-union-address data set}
\label{fig:PLK}
\end{figure} 

\begin{figure}[]
\centering
{\includegraphics[scale = .6]{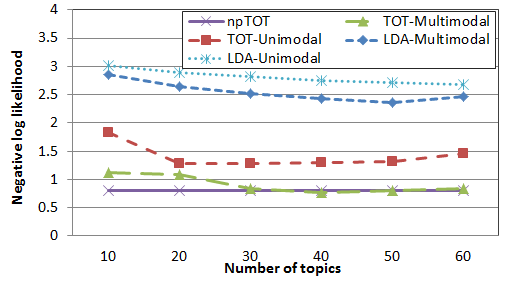}}
\caption{Average per-token negative log likelihood on test set for NIPS dataset.}
\label{fig:NLK}
\end{figure}

Figures \ref{fig:TLk}, \ref{fig:PLK} and \ref{fig:NLK} show the
resulting log joint likelihoods on the three data sets. In each case,
npTOT gives the best likelihood, and the baseline LDA-Unimodal and
LDA-Multimodal models perform poorly. The two TOT models,
TOT-Multimodal  and TOT-Unimodal, perform comparably, and approach the
performance of npTOT as the number of topics reaches that found by
npTOT. This is not surprising; a parametric model with the ``right''
number of topics should perform as well as a nonparametric model. The
advantage of a nonparametric model such as npTOT is that we do not
need to specify the number of topics a priori, or perform expensive
model comparisons, to obtain good results.

\begin{figure}[]
\centering
\includegraphics[scale = .6]{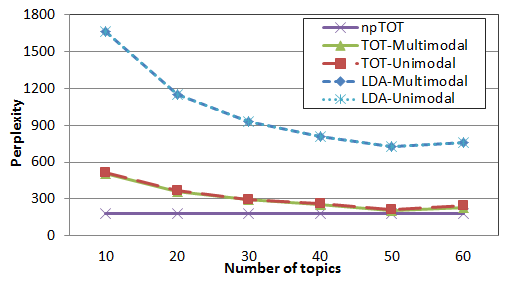}
\caption{Perplexity on test set for Twitter dataset}
\label{fig:TPP}
\end{figure}

\begin{figure}[]
\centering
\includegraphics[scale = .6]{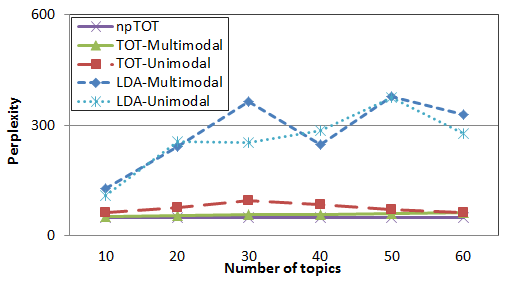}
\caption{Perplexity on test set for State-of-the-union-address data set}
\label{fig:PPP}
\end{figure}

\begin{figure}[]
\centering
\includegraphics[scale = .6]{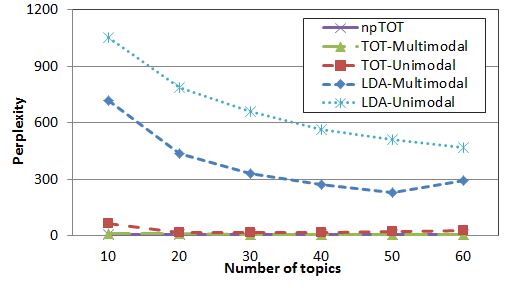}
\caption{Perplexity on test set for NIPS dataset.}
\label{fig:NPP}
\end{figure}

Figures \ref{fig:TPP}, \ref{fig:PPP} and \ref{fig:NPP} show the perplexity obtained through document
completion. Again, we find that npTOT obtains lower perplexity,
indicating that it is better able to predict held-out text. In
particular, note that the LDA models learned without an explicit model
of time perform very poorly. As
expected, having information about when a document is written, and
having a model sophisticated enough to make use of this information,
allows us to make better guesses about the content of that document.



%
%
\section{Conclusions and Future Work}
\label{sec:conclusions}
The goal of this paper was to develop a flexible model for capturing time-varying topics in text corpora where the total number of topics is not known a priori. By extending the TOT model to incorporate nonparametric distributions over both words and timestamps, we have presented a model that is able to find interpretable topics and achieve good predictive performance on held-out data.

One advantage of the npTOT model described herein is that it can easily be extended to higher dimensional covariate values. This would enable us to model geographical variations in topic popularity. In addition to modeling documents, topic models have been used to model images \citep{FeiFei:Perona:2005}. This is another area where spatially dependent topic models, based on the npTOT, could be employed.

While an infinite mixture of Gaussians is flexible and tractable, other distributions may be more applicable in certain situations. For example, when modeling news articles, a topic often peaks suddenly and then dies down gradually. An interesting future approach would be to use asymmetric distributions, such as exponential distributions, to capture this effect.

%
\bibliographystyle{apalike}
\bibliography{ml_project_bib}
\end{document}